\documentclass{article}

\usepackage{arxiv}

\usepackage[utf8]{inputenc} 
\usepackage[T1]{fontenc}    
\usepackage{hyperref}       
\usepackage{url}            
\usepackage{booktabs}       
\usepackage{amsfonts}       
\usepackage{nicefrac}       
\usepackage{microtype}      
\usepackage{lipsum}
\usepackage{booktabs}
\usepackage{floatrow}
\usepackage{amssymb}

\usepackage{amsmath}
\usepackage{listings}

\usepackage{graphicx}

\graphicspath{ {images/} }

\title{Evaluation Framework for large-scale federated learning}

\author{
  Lifeng Liu\\
  Zhejiang University\\
  China, Zhejiang \\
  \texttt{liu\_lf@zju.edu.cn} \\
   \And
 Fengda Zhang \\
  Zhejiang University\\
  China, Zhejiang \\
  \texttt{fdzhang@gmail.com} \\
  \AND
   Jun Xiao\\
  Zhejiang University\\
  China, Zhejiang \\
  \texttt{junx@cs.zju.edu.cn} \\
  \And
    Chao Wu\thanks{Corresponding author: chao.wu@zju.edu.cn} \\
  Zhejiang University\\
  China, Zhejiang\\
  \texttt{chao.wu@zju.edu.cn}\\
}

\begin{document}
\maketitle

\begin{abstract}
Federated learning is proposed as a machine learning setting to enable distributed edge devices, such as mobile phones, to collaboratively learn a shared prediction model while keeping all the training data on device, which can not only take full advantage of data distributed across millions of nodes to train a good model but also protect data privacy. However, learning in scenario above poses new challenges. In fact, data across a massive number of unreliable devices is likely to be non-IID (identically and independently distributed), which may make the performance of models trained by federated learning unstable. In this paper, we introduce a framework designed for large-scale federated learning which consists of approaches to generating dataset and modular evaluation framework. Firstly, we construct a suite of open-source non-IID datasets by providing three respects including covariate shift, prior probability shift, and concept shift, which are grounded in real-world assumptions. In addition, we design several rigorous evaluation metrics including the number of network nodes, the size of datasets, the number of communication rounds and communication resources etc. Finally, we present an open-source benchmark for large-scale federated learning research.

\end{abstract} 


\section{Introduction}
In the present, remarkable achievements have been made in deep learning. Since 2006, the increase in the number and diversity of datasets has been a key factor in the breakthrough of deep learning (the third wave of artificial intelligence). Based on ImageNet \cite{deng2009imagenet}, Alexnet\cite{krizhevsky2012imagenet}, by the construction of the CNN network, has made amazing achievements than ever before, which further proves the importance of good datasets for deep learning models. As people pay more and more attention to data privacy, federated learning \cite{mcmahan2016communication} is proposed to train models with decentralized data while keeping data in devices. The growing demand for federated learning technology has resulted in a lot of algorithms becoming available. Training models via federated learning algorithm, however, still brings some challenges:
 
\textbf{Non-IID Dataset}: Benchmark dataset promotes the development of model training as a common tool for evaluating performace of model. For example, ImageNet, a large-scale and well-structured image dataset, is a milestone which significantly accelerates the advancement of deep convolutional neural networks\cite{simonyan2014very}. One basic hypothesis of machine learning models is that the training and test data should consist of In-dependent and Identically Distributed (I.I.D.) samples. Nevertheless, such property can hardly be guaranteed in practice (especially in federated learning). At present, it is hard to estimate the distribution of these non-IID data with mathematical equations. Moreover, the dataset that can well support the research on non-IID\cite{he2019towards} federated learning is still in vacancy.
 
\textbf{Evaluation Metrics}: Evaluation metrics which is used to assess the performance of models play an important role in research of machine learning. Different machine learning tasks have different evaluation metrics. In traditional centralized deep learning, evaluation metrics mainly include accuracy, precision and recall etc. Accuracy is defined as the proportion of the correct samples to the all samples while evaluating the trained model over test dataset, such as \cite{caldas2018leaf}. Precision (also called positive predictive value) is the fraction of true positive samples among the positive samples, and recall (also known as sensitivity) is the fraction of true positive samples among the samples which are true positive or false negative. In setting of federated learning, the evaluation metrics above are not enough to access the performance of models. For example, the imbalance of heterogeneous data across large-scale nodes, the limitation of storage, computation and communication capacities which are the key factors to the models also need to be considered somehow. 

In this work, we construct and publish well-designed datasets that are delicately designed for supporting non-IID large-scale federated learning. Essentially, we study and explore how to quantitatively describe the distribution of the non-IID dataset\cite{clauset2009power} and the impact of distribution on training models. Besides, we propose an algorithm to modify the existing original classical datasets into non-IID datasets. There are three methods: covariate shift is to skew feature distribution by partitioning the dataset randomly, prior probability shift is to skew label distribution by partitioning the dataset with labels , concept shift\cite{kairouz2019advances} is same feature and different labels by redefining the labels of the dataset with the main concept and context of the dataset. Based on datasets, we introduce multiple evaluation metrics to constitute a complete evaluation metrics of the federated scenario. Except for the accuracy, the framework takes the number of network nodes, the size of datasets, the number of communication rounds, communication resources into account and estimates the performance of different FL algorithms with different tables. Finally, the benchmark data, simulating the federated learning environment, generate from a non-IID dataset and evaluation metrics based on the above mentioned. 

\begin{figure}[h]
    \centering
    \includegraphics[width=0.9\textwidth]{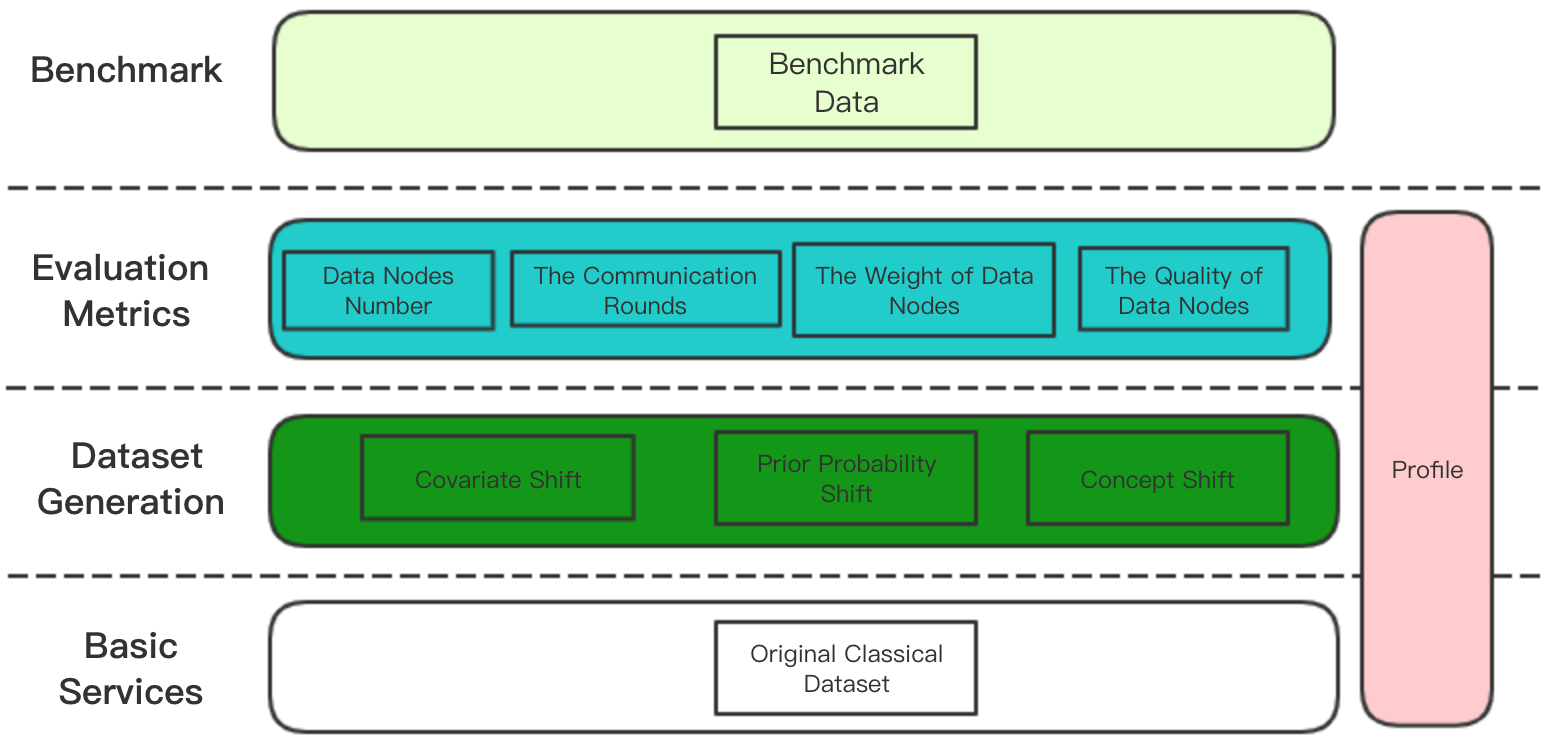}
    \caption{Evaluation Framework Overview}
    \label{fig:structure2}
\end{figure}

To implement the above modular benchmark framework, we show a glimpse of our platform in Figure \ref{fig:structure2}. The platform we called EFFL presented consists of, from bottom to top, basic services module, dataset generation, evaluation metrics, profile and benchmark. The details of the platform implementation will be discussed in the next section.

\textbf{Basic Services}: Basic services are mainly applied to download public original classical data, but also to obtain real-world data collected.

\textbf{Dataset Generation}: In order to facilitate reproducibility, this module is designed for generating Non-IID datasets. The detail information will be shown in Section 2.

\textbf{Profile}: The profile module is designed for configuring different parameters of different datasets to meet the diversified needs. Through this module, we accurately control the generation of the non-IID dataset by setting related parameters.

\textbf{Evaluation Metrics and Benchmark}: In this module, more precise evaluation metrics will be considered to develop a complete evaluation system, the more detail will be shown in Section 3.

\section{Non-IID Dataset}

In traditional neural network learning\cite{bochinski2016training}, the model minimizing empirical error on training data performs well on test data. Unfortunately, in the network of large-scale federated learning, the number and distribution of datasets typically vary significantly across the data holders which challenges the performance of traditional models. Thus, we constructed an open-source, extensible and complete the non-IID dataset that is designed for capturing the intricacies of practical federated environments and promotes research of robust large-scale federated learning methods. As shown in Figure \ref{fig:structure2}, we now detail the non-IID dataset modular.

\subsection{Non-IID Equation Definition}
First of all, we prefer to accurately and intuitively quantify the degree of distribution shift each dataset in the federated learning environment. Based on NI \cite{he2019towards}, we found that different datasets correspond to different trained feature extractors $g_{\varphi}(\cdot)$ and classifiers $f_{\theta}(\cdot)$, and get the corresponding results in the NI equation. In the actual case of federal learning, we found that this equation is not practical and explored a simpler and clearer model to replace the model in the equation. Our goal is to be able to describe the distribution of datasets accurately on a unified scale. Therefore, we replace the feature extractors $g_{\varphi}(\cdot)$ with the fixed Encoder in AutoEncoder model and re-define the Non-IID Index as follow:
\paragraph{Definition 1. Non-IID Encoder Index(NEI)} Given a feature extractor $En(\cdot)$ and a class $C$, the equation is: 
\begin{equation}
NEI(C)=\left\|\frac{\overline{En\left(X_{\operatorname{train}}^{C}\right)}-\overline{En\left(X_{t e s t}^{C}\right)}}{\sigma\left(En\left(X^{C}\right)\right)}\right\|_{2}
\end{equation}
where $X^{C}=X_{train}^{C} \cup X_{test}^{C}$, $\overline{(\cdot)}$ represents the first order moment, $\sigma(\cdot)$ is the std used to normalize the scale of features and $||\cdot||$ represents the 2-norm.

\begin{table}[h!]
\centering
\begin{tabular}{@{}lllll@{}}
\toprule
    & 30\% & 50\% & 70\% & 90\% \\ \midrule
\begin{tabular}[c]{@{}l@{}}Covariate Shift\end{tabular}      & 2.0 & 2.4  & 2.8  & 3.1  \\
\begin{tabular}[c]{@{}l@{}}Prior Probability Shift\end{tabular} & 4.0 & 4.5  & 4.9  & 5.0  \\
\begin{tabular}[c]{@{}l@{}}Concept Shift\end{tabular} & 4.1 & 4.7  & 5.0  & 5.3  \\ 
\bottomrule
\end{tabular}
\caption{The NEI Values on CIFAR10}
\label{table:NEI}
\end{table}

As illustrated in Table \ref{table:NEI}, the first row represents the proportion of the partitioned dataset to the source dataset. In our experiments, the fixed entire Encoder looks like Conv->MaxPool->Conv->Upsample->Conv->Conv. The MaxPool layer above is replaced with a Conv with num channels=32, kernel size=3 and strides=2. With the partitioned dataset became larger, the NEI value is higher. The NEI values in partitioning randomly method is the smallest. Opposite, it's higher in the redefining digit labels method. The showcase and statistical analysis well support an plausible conclusion that the degree of distribution shift quantified by NEI is a key factor influencing classification performance. 

\subsection{Non-IID Dataset Generation}
In this part, we focus to design datasets that are sufficient to approximate data in the federated scenarios. Naturally, collecting data in a real federated environment is the best choice. Besides, The existing classic dataset has been of great developed. After transformation with the federated method, we design it into the dataset we need. Therefore, there are three ways of partitioning them. All of them can be evaluated by the NEI. The workflow are in Figure \ref{fig:datageneration}:

\begin{figure}[h]
    \centering
    \includegraphics[width=0.8\textwidth]{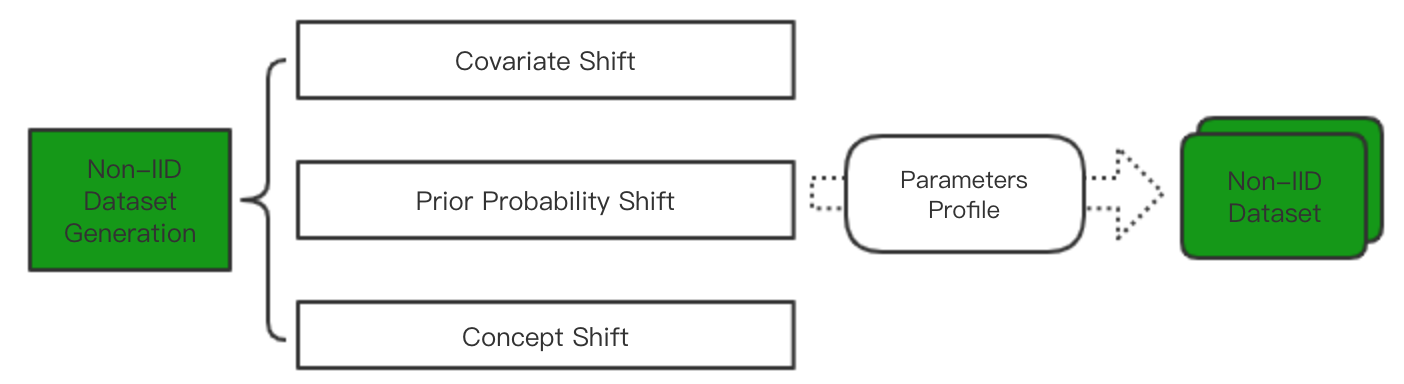}
    \caption{Non-IID Dataset Generation Overview}
    \label{fig:datageneration}
\end{figure}

As illustrated in Figure \ref{fig:datageneration}, the specific steps are as follows:
\begin{enumerate}
  \item Selecting the corresponding dataset partition methods.
  \item Setting parameters profile including datasets, the number of nodes, etc.
  \item Based on the parameters profile (profile module), generate Non-IID datasets. 
\end{enumerate}

\subsubsection{Covariate Shift}
In the federated learning scene, it is common that the same label corresponds to training datasets with different features. Covariate shift means feature distribution skews\cite{kairouz2019advances}. The marginal distribution $\mathcal{P}_{i}(x)$ may vary cross clients, even if $\mathcal{P}(x|y)$ is shared. Because of $\mathcal{P}_{i}(x|y)= \mathcal{P}_{j}(x|y)$ for all client $i$ and $j$, we will add different parameters of noise such as Gaussian noise or salt and pepper noise to the training data of the dataset in different clients to simulate $\mathcal{P}_{i}(x)$ varies widely. 

%
\subsubsection{Prior Probability Shift}
Facing the complex distributed data challenge in federated learning, we found that simply generating a fixed dataset does not meet the complexity requirements of the current federated learning environment. Prior probability shift is proposed to describe this scenario, which is defined as the marginal distributions $\mathcal{P}(y)$ may vary cross clients, even if $\mathcal{P}(x|y)$ is the same. Thus, in this methodology, we explore a mature, dynamically configurable solution scheme and reproduce results. For federated environments with different NEI values, we dynamically configure a series of related parameters, such as the number of nodes, datasets size, etc.

For example, when we partition the data according to the label, we assume that each dataset and other datasets are non-IID distribution. In reality, these datasets are more complex. Some of them are the same or their distribution may be partially the same, or there are some "dirty" data (error data), or noise data is added, etc. After a thorough investigation, we have established a suite of open source and extensible parameter setting profile.

\subsubsection{Concept Shift}
In addition, it is common for labels that reflect different sentiment and different meanings of an image, etc. Concept Shift, same feature and different label, proposed to describe the conditional distribution $\mathcal{P}_{i}(x|y)$ may vary cross clients, even if $\mathcal{P}_{i}$ is same.

Based on the \cite{he2019towards}, we will represent the label of dataset image in two dimensions: main concept and contexts which provides a novel perspective on the classification of dataset images. For example, in CIFAR10 \cite{krizhevsky2010convolutional} dataset, in the category of ‘dog’, images are divided into different contexts such as ‘grass’, ‘car’, ‘beach’, meaning the ‘dog’ is on the grass, in the car, or on the beach respectively. With these contexts, one can easily design an Non-IID setting. Naturally, in the category of 'grass' - the same context, images are divided into different concepts such as 'dog', 'cat', etc. Meanwhile, the degree of distribution shift can be flexibly controlled by adjusting the proportions of different contexts.

\subsubsection{Unbalancedness}
For the IID dataset, where the data is shuffled, we partition the datasets randomly and get multiple shards for unbalancedness. Each small shards represents independent data holders. Unlike the normal partition, we set the size of every shard and investigate the effect of different data size on model performance.

\section{Evaluation Metrics and Benchmark}
Rigorous evaluation metrics are required to appropriately assess how a learning solution behaves in federated scenarios. However, at present, there is no convincing standard general evaluation metrics. In this section, we hope to establish an initial set of metrics chosen specifically for this purpose. Except for the accuracy, we propose four general evaluation metrics: the number of data nodes, the communication round, the weight of data nodes\cite{torralba2011unbiased} and the data quality of data nodes. All the experiments are based on the server with 128G memory, 128-core Intel Xeon CPUs and two v100 NVIDIA GPUs.

\textbf{Data Nodes Number}: In federated learning, it is an unavoidable learning process that multiple nodes participate in learning to get a global model. Naturally, the number of training participants in each training round is a key factor for the performance of the model. We compare the experimental results of the traditional FedAvg algorithm in which both of the communication rounds is 2000 , the same way of initial weight node and N\verb|\|E is 0\% . 

\begin{table}[h!]
    \centering
    \begin{tabular}{@{}lllll@{}}
    \toprule
    On MNIST & 5 nodes & 10 nodes & 20 nodes & 30 nodes \\ 
    \midrule
    \begin{tabular}[c]{@{}l@{}}Quantity Skew\end{tabular}      
    & 92.55\% & 94.32\%  & 96.83\%  & 97.21\%  \\
    \begin{tabular}[c]{@{}l@{}}Label Distrubition Skew\end{tabular} 
    & 80.07\% & 83.11\%  & 88.36\%  & 93.22\%  \\ 
    \bottomrule
    \toprule
    On CIFAR10 & 5 nodes & 10 nodes & 20 nodes & 30 nodes \\ 
    \midrule
    \begin{tabular}[c]{@{}l@{}}Quantity Skew\end{tabular}      
    & 91.83\% & 91.92\%  & 95.94\%  & 94.33\%  \\
    \begin{tabular}[c]{@{}l@{}}Label Distrubition Skew\end{tabular} 
    & 68.10\% & 69.43\%  & 71.86\%  & 71.50\%  \\ 
    \bottomrule
    \end{tabular}
    \caption{Data Nodes Number Results}
    \label{table:23}
\end{table}

As illustrated in Table \ref{table:23}, the final results show that in a certain range, the more nodes participate in the training at the same time, the better the training effect. What's more, with the same node nums, the performance of method on MNIST\cite{lecun1995learning} is better than on CIFAR10.

\textbf{The Communication Rounds}: There is no doubt that the communication rounds of nodes play an important role in the performance of the model. Due to the uncertainty of the federated network, communication is huge resource consumption. We hope to decrease the number of communication rounds and improve the accuracy of training. Naturally, except the communication rounds, the other factors is same.

\begin{table}[h!]
    \centering
    \begin{tabular}{@{}llllll@{}}
    \toprule
    On MNIST & 500     & 1000    & 1500    & 2000    & 3000    \\ 
    \midrule
    \begin{tabular}[c]{@{}l@{}}Quantity Skew\end{tabular}      
    & 80.33\% & 85.41\% & 89.55\% & 94.77\% & 96.33\% \\
    \begin{tabular}[c]{@{}l@{}}Label Distrubition Skew\end{tabular} 
    & 80.32\% & 81.58\% & 86.53\% & 89.55\% &93.44\% \\
    \midrule
    \toprule
    On CIFAR10 & 500    & 1000    & 1500    & 2000    & 3000    \\
    \midrule
    \begin{tabular}[c]{@{}l@{}}Quantity Skew\end{tabular}      
    & 74.11\% & 86.91\% & 88.56\% & 95.08\% & 95.94\% \\
    \begin{tabular}[c]{@{}l@{}}Label Distrubition Skew\end{tabular} 
    & 65.10\% & 66.53\% & 68.86\% & 70.44\% & 71.50\% \\
    \bottomrule
    \end{tabular}
    \caption{The Communication Rounds Results}
    \label{table:CommunicationRoundsResults}
\end{table}

As illustrated in Table \ref{table:CommunicationRoundsResults}, obviously, under the same conditions, with the increase of communication rounds, the global model training curve is to converge.

\textbf{The Weight of Data Nodes}: We need to recognizes the importance of specifying how the accuracy is weighted across nodes, e.g., whether every node is equally important, or every data node equally important (implying that the more data, the more important the node).

\textbf{The Quality of Data Nodes}: In this part, we research the influence of the same distributed data or the same data proportion in the total data and the wrong data proportion in the total data proportion on the performance of the model. We regard the above properties as the quality of data. Except the quality of data nodes, the other factors is no different.

\begin{table}[h!]
\centering
    \begin{tabular}{@{}lllllll@{}}
    \toprule
    On MNIST & \multicolumn{3}{l}{Quantity Skew} & \multicolumn{3}{l}{Label Distrubition Skew} \\ \midrule
N\textbackslash{}E & 0\% & 5\% & 10\% & 0\% & 5\% & 10\% \\
0\% & 96.54\% & 95.33\% & 92.90\% & 93.97\% & 90.33\% & 85.21\% \\
10\% & 96.31\% & 92.90\% & 88.44\% & 92.83\% & 89.40\% & 86.33\%\\
20\% & 96.90\% & 95.44\% & 93.15\% & 92.05\% & 90.14\% & 88.50\%\\
30\% & 96.20\% & 92.78\% & 90.32\% & 90.33\% & 87.33\% & 84.33\%\\
\bottomrule
\toprule
    On CIFAR10 & \multicolumn{3}{l}{Quantity Skew} & \multicolumn{3}{l}{Label Distrubition Skew} \\ \midrule
N\textbackslash{}E & 0\% & 5\% & 10\% & 0\% & 5\% & 10\% \\
0\% & 94.1\% & 92.23\% & 90.23\% & 72.41\% & 69.83\% & 68.42\% \\
10\% & 95.33\% & 93.76\% & 91.62\% & 73.30\% & 71.73\% & 68.04\%\\
20\% & 95.28\% & 94.11\% & 92.33\% & 75.66\% & 73.92\% & 71.48\%\\
30\% & 95.54\% & 93.67\% & 92.74\% & 78.53\% & 75.31\% & 72.43\%\\
\bottomrule
\end{tabular}
\caption{The Quality of Data Nodes Results}
\label{table:Partitioning1}
\end{table}

As illustrated in Table \ref{table:Partitioning1}, N represents the proportion of the same dataset in total dataset. E represents the proportion of the error data in total dataset. In a certain range, under the same conditions, the larger N values is, the higher the accuracy of the model is, and the larger E is, the lower the accuracy of the model is. 

\textbf{The tutorial of the evaluation framework}:
At present, the classification mentioned above can include all datasets. The specific implementation method is currently mainly for MNIST and cifar10 datasets. Moreover, we open source our code and some non-IID datasets\footnote{https://github.com/ZJU-DistributedAI/DAIDataset}. 

To generate Non-IID datasets, we need to:
\begin{enumerate}
    \item install Python 3.6 environment 
    \item run command line: pip3 install -r requirements.txt
\end{enumerate}

In our framework, we create the config.yaml to register the function of profile. In downloadDataset.py, the classical datasets will be downloaded. we replace dataset gerneration module with makeDataset module and preprocess module, etc. 
The workflow is following:

\begin{enumerate}
  \item Editing the config.yaml according to request such as dataset\_mode, node\_num and etc.
   \begin{lstlisting}
    # dataset
    dataset_mode: CIFAR10
    
    # node num Number of node(default n=20) one node corresponding to one dataset
    node_num: 10
    
    # partition methods, dataset partition, 0-covariate shift, 1-prior propability shift, 2-concept shift
    split_mode: 0
    \end{lstlisting}
  \item Selecting the method of data generation.
  \item Execute the downloadData module(downloadData.py).
  \item Estimate the NEI values in NEI module: python3 NEI.py
  \item Execute makeDataset and preprocess modules.
    Run the program: python3 makeData.py, the terminal will show:
     \begin{lstlisting}
    begin
    .......
    index 5 saved
    saved file succeed !
    \end{lstlisting}
    
  \item Test the partitioned non-IID dataset.
  \item Generate a benchmark data.
    \begin{figure}[h]
        \centering
        \includegraphics[width=0.6\textwidth]{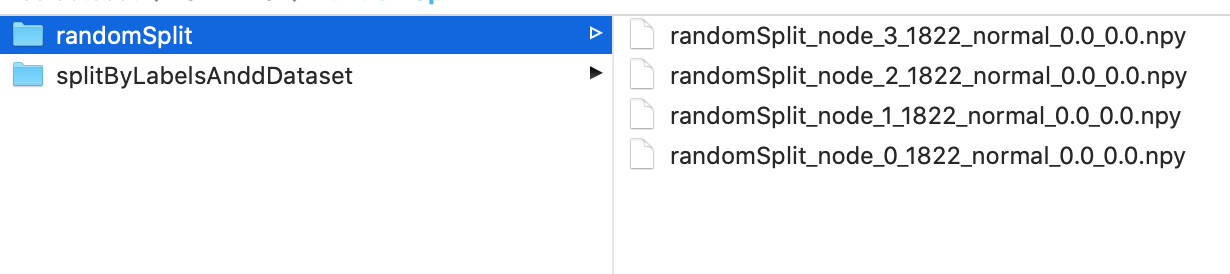}
        \caption{The non-IID Datasets Overview}
        \label{fig:editconfig}
    \end{figure}
\end{enumerate}
The more detail information in link\footnote{https://zju-distributedai.github.io/GalaxyDataset/docs/quick-start-guide/}.

\section{Conclusions and Future Works}
In this paper, we introduce a novel evaluation framework for large-scale federated learning. We present a complete, scalable, open-source non-IID dataset. Moreover, a suite of evaluation metrics is proposed as a framework to evaluate the performance of federated learning algorithms. Finally, we release a benchmark result for the related research. 

We will focus on the following works. Firstly, we still need to explore more simple NEI equations to evaluate the non-IID dataset. What's more, we will explore more practical and effective dataset partitioning methods. Secondly, more settings about different forms of Non-IID are expected to be exploited.







\bibliographystyle{ieeetr} 
\bibliography{references} 

\end{document}